\documentclass[letterpaper,journal]{IEEEtran}
\usepackage{amsmath,amsfonts}
\usepackage{algorithmic}
\usepackage{algorithm}
\usepackage{array}
\usepackage[caption=false,font=normalsize,labelfont=sf,textfont=sf]{subfig}
\usepackage{textcomp}
\usepackage{stfloats}
\usepackage{url}
\usepackage{verbatim}
\usepackage{graphicx}
\usepackage{cite}
\usepackage{multirow}
\makeatletter
\begin{document}

\title{Dense-depth map guided deep Lidar-Visual Odometry with Sparse Point Clouds and Images}

\author
{Junying Huang,  Ao Xu, Dongyong Sun\\
Yuanfeng Wang, and Qi Qin
\thanks{The work was supported by Guangdong Provincial Quantum
Science Strategic Initiative (GDZX2306001) and the startup fund of Shenzhen City. (Corresponding author: Qi Qin.)}
\thanks{Junying Huang is with the College of Physics and Optical Engineering, Shenzhen University, Shenzhen 518060, China (e-mail: 2210452118@email.szu.edu.cn).}
\thanks{Ao Xu is with the Research Institute of Tsinghua University in Shenzhen, Shenzhen 518057, China. (e-mail: ao.xu@foxmail.com).}
\thanks{Dongyong Sun and Zixiang Wang are with YunJiZhiHui Engineering Co., Ltd., Shenzhen 518049, China (e-mail: dysun.sunny@gmail.com; 2401762018@qq.com).}
\thanks{Yuanfeng Wang is with the Quantum Science Center of the Guangdong-Hong Kong-Macao Greater Bay Area, Shenzhen 518045, China (e-mail: wangyuanfeng@quantumsc.cn).}
\thanks{Qi Qin is with College of Physics and Optical Engineering, Shenzhen University, Shenzhen 518060, China, and also with State Key Laboratory of Radio Frequency Heterogeneous Integration, Shenzhen University, Shenzhen 518060, China, and also with Institute of Intelligent Optical Measurement and Detection, Shenzhen University, Shenzhen 518060, China, and also with Quantum Science Center of Guangdong-HongKong-Macao Greater Bay Area (Guangdong), Shenzhen 518045, China (e-mail: qi.qin@szu.edu.cn). }}

\markboth{Journal of \LaTeX\ Class Files,~Vol.~14, No.~8, August~2021}%
{Shell \MakeLowercase{\textit{et al.}}: A Sample Article Using IEEEtran.cls for IEEE Journals}


\maketitle

\begin{abstract}
Odometry is a critical task for autonomous systems for self-localization and navigation. We propose a novel LiDAR-Visual odometry framework that integrates LiDAR point clouds and images for accurate and robust pose estimation. Our method utilizes a dense-depth map estimated from point clouds and images through depth completion, and incorporates a multi-scale feature extraction network with attention mechanisms, enabling adaptive depth-aware representations. Furthermore, we leverage dense depth information to refine flow estimation and mitigate errors in occlusion-prone regions. Our hierarchical pose refinement module optimizes motion estimation progressively, ensuring robust predictions against dynamic environments and scale ambiguities. Comprehensive experiments on the KITTI odometry benchmark demonstrate that our approach achieves similar or superior accuracy and robustness compared to state-of-the-art visual and LiDAR odometry methods. 
\end{abstract}


\section{Introduction}
\IEEEPARstart{O}{dometry} estimates a robot or vehicle’s pose using sensor data such as IMU, camera, or LiDAR, and is essential in robotics, autonomous driving, and AR applications \cite{shan2020lio, zhang2015visual, li2021self}. Visual Odometry (VO) relies on RGB images, which offer rich texture but suffer from depth ambiguity, lighting variation, and occlusion \cite{tateno2017cnn, sun2018pwc}. LiDAR Odometry (LO) uses 3D point clouds for geometric and depth information and performs well in pose estimation \cite{du2021lidar, zhang2014loam, shan2018lego}, but is limited by sparse data, sensor noise, and environmental sensitivity.

LiDAR-Visual Odometry (LVO) fuses both modalities to improve robustness and accuracy \cite{geiger2013vision, xu2019depth}. In this paper, we propose an LVO framework that combines point clouds and RGB images, and integrates dense-depth maps to enhance depth representation across the pipeline. Specifically, we estimate dense-depth maps from sparse LiDAR and images to overcome occlusions and noise, and fuse them with RGB to form a four-channel input for end-to-end pose estimation. This improves the reliability of depth features critical for motion estimation.

The main contributions are:
\begin{itemize}
\item A novel dense-depth guided LVO method that fuses LiDAR and visual features to improve depth representation.
\item Use of spatial, channel, and cross attention mechanisms to effectively extract and fuse multimodal features \cite{vaswani2017attention}.
\item A depth-aware optical flow module with hierarchical refinement to improve flow estimation \cite{sun2018pwc}.
\item Depth-integrated pose refinement, enhancing translation estimation accuracy.
\end{itemize}

We validate our method on the KITTI dataset \cite{geiger2012we}, showing that it outperforms state-of-the-art LVO approaches on most sequences in terms of accuracy and robustness.

The rest of the paper is structured as follows: Section II reviews related works; Section III describes the proposed method; Section IV presents experiments; Section V concludes the paper.

\section{Related Work}
Robust pose estimation is key to autonomous navigation. Traditional methods like LOAM \cite{zhang2014loam}, LeGO-LOAM \cite{shan2018lego}, and LIO-SAM \cite{shan2020lio} rely on handcrafted features and geometric constraints but face challenges like drift and computational cost. Recently, deep learning-based odometry methods have gained traction for their strong feature learning ability. We review advances in deep VO, LO, and VLO.

\subsection{Deep Visual Odometry}
Classical VO methods use feature-based \cite{mur2015orb, mur2017orb, engel2017direct} or direct techniques \cite{newcombe2011dtam, engel2014lsd}, but suffer under poor lighting, low texture, and occlusion. Deep learning approaches address these limitations. DeepVO \cite{wang2017deepvo} introduced an RNN to model temporal dependencies. DF-VO \cite{zhan2021df} employed self-supervised learning with feature matching and uncertainty modeling. RAFT-SLAM \cite{teed2020raft} improved flow estimation with all-pairs field transforms. However, monocular VO lacks scale and struggles in dynamic or fast-motion scenes.

To address scale ambiguity, stereo-based models like DeepStereoVO \cite{zhan2018unsupervised} and D3VO \cite{yang2020d3vo} were proposed, incorporating stereo constraints and uncertainty-aware depth. Still, VO remains vulnerable to illumination and texture variations.

\subsection{Deep LiDAR Odometry}
LO estimates motion from point clouds using ICP \cite{besl1992method} or feature-based methods \cite{zhang2014loam, shan2020lio, ye2019tightly}, but suffers from noise and sparsity. Learning-based LO methods learn features directly from LiDAR. LO-Net \cite{li2019net} extracts geometric features with a deep network. DeepLO \cite{cho2019deeplo} fuses learning with ICP refinement. PWCLO \cite{wang2021pwclo} uses point-wise correlations for efficient real-time estimation. LodoNet \cite{zheng2020lodonet} improves accuracy via 2D keypoint matching on projected LiDAR data.

\subsection{Visual-LiDAR Odometry (VLO)}
VLO leverages complementary RGB and LiDAR data for improved robustness \cite{zhang2015visual, maddern2014illumination}. Deep learning-based VLO models have shown strong performance. DVLO \cite{liu2024dvlo} employs local-global feature fusion with bi-directional alignment. An et al. \cite{an2022visual} proposed an unsupervised multi-channel network for accurate mapping and localization in dynamic scenes.

While recent methods improve accuracy, challenges remain in real-time inference, dynamic environments, and cross-modal calibration. Our method builds on these by integrating depth completion and attention mechanisms to boost pose estimation performance.

\begin{figure*}[t]
    \centering
    \includegraphics[width=1.0\textwidth]{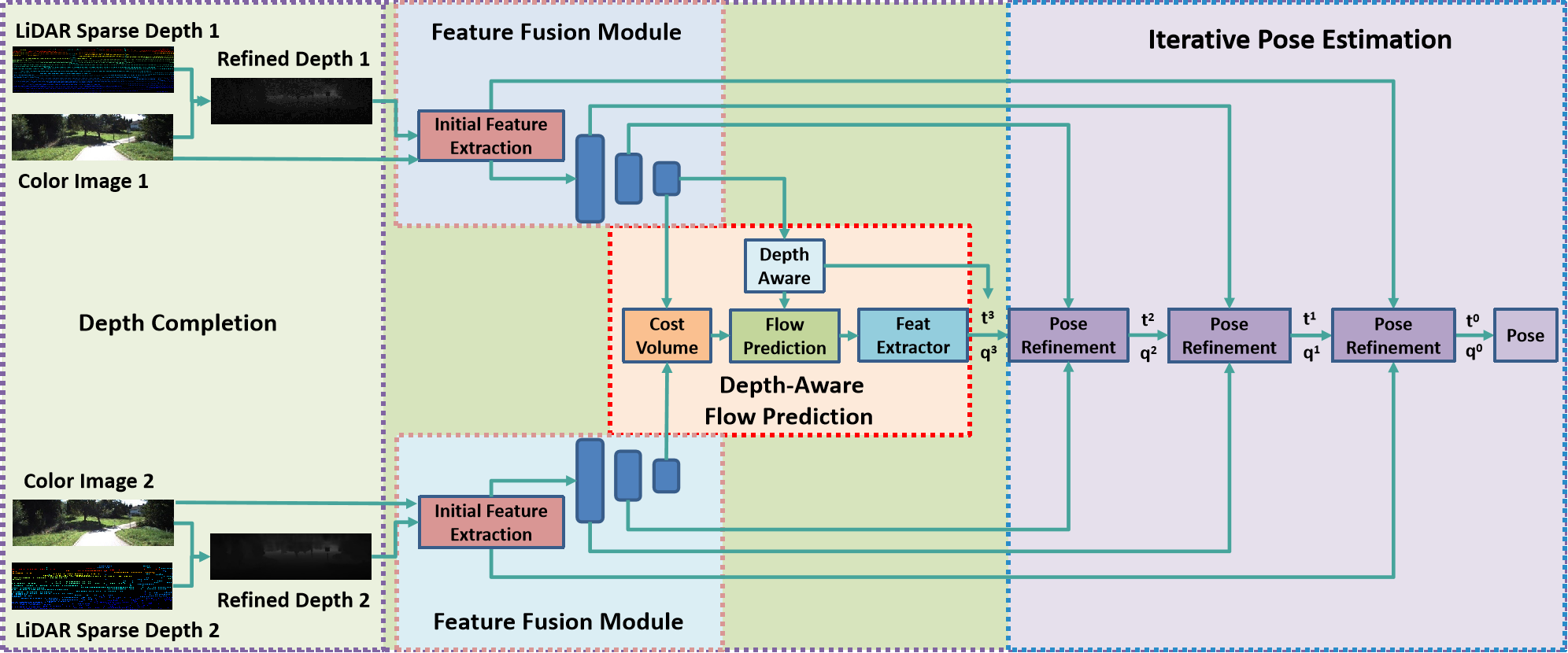}
    \caption{Overview of our proposed D3LVO framework. The network consists of depth completion, multi-scale feature extraction with attention mechanisms, depth-aware optical flow prediction, cost volume computation, and hierarchical pose refinement modules.}
    \label{fig:network_architecture}
\end{figure*}

\section{Methodology}
In this section, we introduce our \textbf{d}ense-\textbf{d}epth map based \textbf{d}eep \textbf{L}idar-\textbf{V}isual \textbf{O}dometry (D3LVO) framework, which integrates LiDAR point clouds and RGB images for robust pose estimation. Our method consists of five key components: depth completion, multi-scale feature extraction, cost volume computation, depth-aware optical flow prediction, and hierarchical pose refinement. The overall architecture is shown in FIGURE~\ref{fig:network_architecture}.

\subsection{Depth Completion}
LiDAR point cloud is often sparse and prone to measurement noise, which pose significant challenge for robust odometry applications. To mitigate these shortcomings and enhance feature integrity, we employ PENet\cite{hu2021penet}, a learning-based depth completion method that integrates geometric priors from the LiDAR scans with structural constraints from the RGB image. PENet refines depth information through a residual learning strategy, generating dense and high-quality depth maps. The enhanced depth map is concatenated with the RGB image to form a four-channel input (RGB-D) for subsequent processing.

\subsection{Multi-Scale Feature Extraction with Attention Mechanisms}
To extract meaningful representations from RGB-D inputs, we employ a hierarchical feature extraction network inspired by PWC-Net \cite{sun2018pwc}. The feature pyramid contains four levels, progressively increasing the receptive field while maintaining fine-grained spatial details. \textbf{Level 0 ($h/2, w/2$)} employs convolutional layers and residual blocks to capture low-level texture and depth cues. \textbf{Level 1 ($h/4, w/4$)} increases feature depth for richer semantic representation. \textbf{Level 2 ($h/8, w/8$)} abstracts motion-related features. Finally, \textbf{Level 3 ($h/16, w/16$)} serves as the coarsest level, providing the initial input for optical flow estimation.

\subsubsection{Attention-Guided Feature Extraction}
To enhance feature quality, we incorporate multiple attention mechanisms that dynamically refine feature maps. \textbf{Channel Attention (CA)} \cite{hu2018squeeze} is applied to RGB features to strengthen discriminative channels using a squeeze-and-excitation structure, expressed as:
\begin{equation}
    \mathbf{CA}(\mathbf{x}) = \sigma \left( w_2 \delta \left( w_1 \left[ \mathit{GAP}(\mathbf{x}), \mathit{GMP}(\mathbf{x}) \right] \right) \right) \odot \mathbf{x},
\end{equation}
where \( \mathit{GAP}(\mathbf{x}) \) and \( \mathit{GMP}(\mathbf{x}) \) represent global average and max pooling, \( w_1 \) and \( w_2 \) denote fully connected layers, \( \delta \) is the ReLU activation, \( \sigma \) is the sigmoid function, and \( \odot \) denotes element-wise multiplication with the input.

\textbf{Spatial Attention (SA)} \cite{woo2018cbam} is applied to the depth stream to emphasize depth-salient regions and enhance geometric perception. It is formulated as:
\begin{equation}
    \mathbf{SA}(\mathbf{x}) = \sigma \left( f^{3 \times 3} \left[ \mathit{MaxPool}(\mathbf{x}), \mathit{AvgPool}(\mathbf{x}) \right] \right) \odot \mathbf{x},
\end{equation}
where \( \mathit{MaxPool}(\cdot) \) and \( \mathit{AvgPool}(\cdot) \) are spatial pooling operations across channels, and \( f^{3 \times 3} \) is a convolution layer that learns spatial attention weights.

To further enhance modality fusion, we adopt \textbf{Cross Attention (XA)} \cite{vaswani2017attention}, which enables mutual interaction between RGB and depth features by attending to complementary information:
\begin{equation}
    \begin{split}
        \mathbf{XA}(\mathbf{x}_r, \mathbf{x}_d) &= \operatorname{SoftMax} \left( \frac{q_r k_d^T}{\sqrt{d_k}} \right) v_d \\
        &\quad + \operatorname{SoftMax} \left( \frac{q_d k_r^T}{\sqrt{d_k}} \right) v_r,
    \end{split}
\end{equation}
where \( q, k, v \) denote the query, key, and value embeddings from the RGB (\( r \)) and depth (\( d \)) branches, and \( d_k \) is a scaling factor to stabilize training.

\begin{figure}[t]
    \centering
    \includegraphics[width=0.42\textwidth]{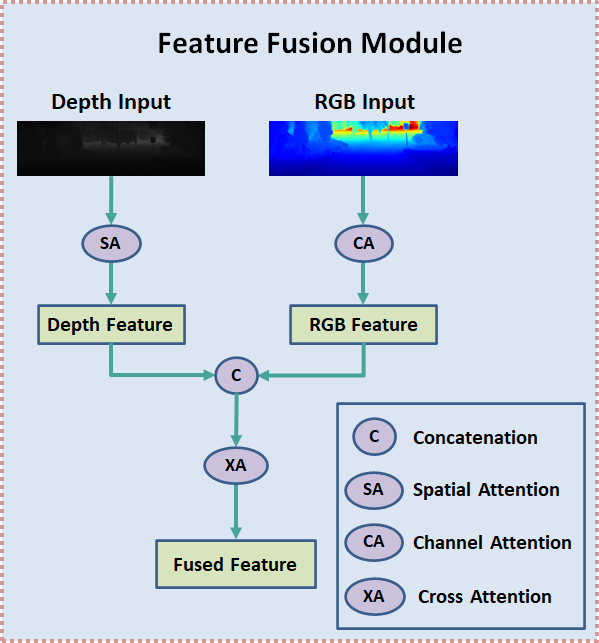}
    \caption{Illustration of our attention-guided feature extraction/fusion module. Channel Attention (CA) enhances RGB features, Spatial Attention (SA) refines depth features, and Cross Attention (XA) facilitates mutual interaction between the two modalities.}
    \label{fig:feature_fusion}
\end{figure}

Through these attention mechanisms, the network is guided to focus on informative regions in both RGB and depth modalities, enabling more robust feature representations that improve both optical flow estimation and pose refinement.

\subsection{Cost Volume Computation}
To estimate the pixel-level motion between two frames, we construct a cost volume at each pyramid level and follow  hierarchical refinement process inspired by the multi-scale approaches for robust optical flow estimation \cite{sun2018pwc,hui2018liteflownet,zhao2020maskflownet}. The cost volume encodes the similarity between feature representations from two consecutive frames, allowing the network to estimate motion robustly. Given two feature maps, $\mathbf{F}_1$ and $\mathbf{F}_2$, extracted from two consecutive frames, the cost volume is computed as:

\begin{equation}
    C(x, y, d_x, d_y) = \sum_{c=1}^{C} \mathbf{F}_1^c(x, y) \cdot \mathbf{F}_2^c(x + d_x, y + d_y),
\end{equation}
where $(x, y)$ represents the pixel coordinates, $c$ indexes the feature channels, and $(d_x, d_y)$ denotes the displacement within the search range $S=4$. The network considers all possible displacements in a predefined search range $S=4$, leading to a cost volume of size $(2S+1)^2 \times H \times W$ for each feature map pair. 

To ensure stability during optimization, we normalize the feature maps before computing the cost volume:
\begin{equation}
    \tilde{\mathbf{F}}_i = \frac{\mathbf{F}_i}{\|\mathbf{F}_i\|_2}, \quad i \in \{1,2\},
\end{equation}
where $\|\cdot\|_2$ denotes the L2 norm. This normalization ensures that the computed similarity scores remain bounded and prevents feature magnitude variations from affecting the correlation response.

The cost volume is constructed by iteratively shifting $\mathbf{F}_2$ over a local search window of size $(2S+1) \times (2S+1) = 9 \times 9$, computing the inner product at each shift. Specifically, for each displacement $(d_x, d_y)$ within the search range, we shift $\mathbf{F}_2$ by $(d_x, d_y)$ and compute the element-wise product with $\mathbf{F}_1$. This operation is implemented efficiently using tensor operations, avoiding redundant computation.
The constructed cost volume is subsequently processed by 2D convolutional layers to extract motion-related features. These convolutional layers consist of multiple cascaded convolutions with ReLU activations, which learn to encode spatial and temporal motion patterns from the cost volume. The extracted features are then passed to the optical flow estimation module, where they are progressively refined at each pyramid level to produce accurate flow predictions.

\subsection{Depth-Aware Optical Flow Prediction}

\begin{figure}[t]
    \centering
    \includegraphics[width=0.40\textwidth]{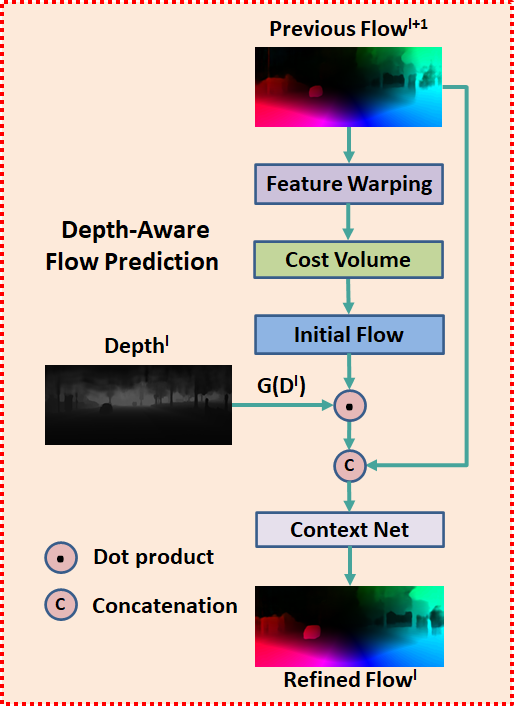}
    \caption{Illustration of the depth-aware optical flow prediction module. The optical flow is visualized using the HSV color coding scheme, where the hue represents the flow direction and the saturation/value represents the flow magnitude. Specifically, red indicates rightward motion, green indicates downward motion, blue indicates leftward motion, and the color intensity corresponds to the motion magnitude.}
    \label{fig:depth_guided_flow}
\end{figure}

Optical flow estimation plays a crucial role in our D3LVO framework. We estimate optical flow hierarchically, refining it progressively from coarse to fine levels. To improve accuracy, particularly in low-texture regions and occlusions, we introduce depth guidance at each level. Unlike previous approaches that directly concatenate depth as an additional input for feature extraction\cite{mittal2020just}, we leverage depth to adaptively scale the predicted flow magnitude while maintaining the structural integrity of the cost volume. This depth modulation helps reduce scale ambiguity and improves flow consistency in textureless and occluded regions, as shown in\cite{dong2023rethinking}. Furthermore, we incorporate geometric constraints to enhance robustness in dynamic environments\cite{bao2019depth}. The depth-aware flow prediction module is shown in FIGURE~\ref{fig:depth_guided_flow}.

\subsubsection{Depth-Aware Flow Estimation}

At each pyramid level, optical flow is estimated from a cost volume computed using feature maps of consecutive frames. The flow prediction is modulated by depth through:

\begin{equation}
    \mathbf{u}_i = \mathcal{F}(\mathbf{F}_1^i, \mathbf{F}_2^i, \mathbf{C}_i) \cdot \mathcal{G}(D_i),
\end{equation}

where $\mathbf{u}_i$ is the flow at level $i$, $\mathbf{F}_1^i$, $\mathbf{F}_2^i$ are feature maps, $\mathbf{C}_i$ the cost volume, and $D_i$ the depth map. The flow estimator $\mathcal{F}(\cdot)$ is a lightweight CNN using multiple \(1 \times 1\) convolutions with batch normalization and Leaky ReLU, designed for efficiency and robustness in real-time settings. The depth modulation function $\mathcal{G}(D_i)$ consists of convolution layers expanding channels to 32, followed by \(1 \times 1\) convolution and activation, producing a depth-aware weight map to scale the flow magnitude. The flow module employs a 3-layer MLP (sizes 128, 64, 2) based on \(1 \times 1\) convolutions, taking as input a concatenation of feature map, cost volume, depth features, and optionally the previous flow processed by FlowNet. This design efficiently captures motion features while maintaining a compact architecture \cite{sun2018pwc, hui2018liteflownet}.

Depth guidance offers several advantages:
\begin{itemize}
    \item \textbf{Scale Ambiguity Reduction}: Adaptive scaling based on depth mitigates monocular scale ambiguity \cite{teed2020raft}.
    \item \textbf{Improved Robustness in Textureless Regions}: Depth provides geometric constraints improving flow accuracy where textures are sparse \cite{bao2019depth}.
    \item \textbf{Occlusion Handling}: Depth-aware weights help differentiate occluded areas, enhancing flow consistency \cite{zhao2020maskflownet}.
\end{itemize}

\subsubsection{Residual Flow Refinement}

To refine flow, a residual learning scheme with depth scaling is applied:

\begin{equation}
    \mathbf{u}_i = \mathbf{u}_{i+1}^{\uparrow} + \mathcal{G}(D_i) \cdot \Delta \mathbf{u}_i,
\end{equation}

where $\mathbf{u}_{i+1}^{\uparrow}$ is the upsampled flow from the coarser level, $\Delta \mathbf{u}_i$ is the residual correction, and $\mathcal{G}(D_i)$ the depth-aware factor. This enables learning fine motion details consistent with scene geometry. A Context Net further improves refinement by combining initial and refined flows. It extracts high-level context from concatenated flow features to guide residual correction, enhancing robustness to large motion and occlusion \cite{hui2018liteflownet}.


\begin{figure}[t]
    \centering
    \includegraphics[width=0.42\textwidth]{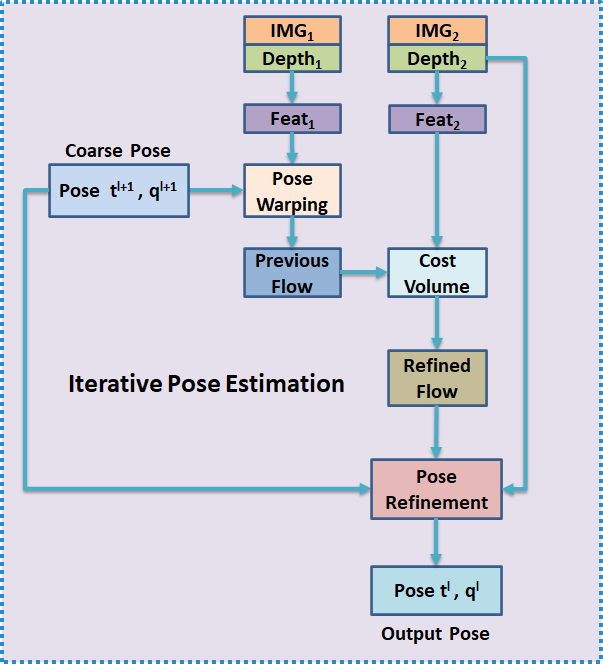}
    \caption{Overview of Depth-Aware Pose Estimation: The proposed Pose Warp-Refinement module at the l-th level fuses multi-scale optical flow features and depth maps to refine pose estimates across hierarchical levels.}
    \label{fig:depth_aware_pose}
\end{figure}

\subsection{Hierarchical Pose Refinement}
Accurate pose estimation is crucial for odometry tasks, especially in visual-inertial and Lidar-based systems\cite{mur2015orb, engel2017direct}. Traditional deep learning-based visual odometry approaches, such as DeepVO\cite{wang2017deepvo}, rely on recurrent structures for sequential motion estimation, while methods like RAFT-SLAM\cite{teed2020raft} leverage all-pairs correlation for dense flow-based pose estimation. In contrast, we propose a hierarchical pose refinement strategy that incorporates depth information to improve scale recovery and translation accuracy.The overall pose refinement process is progressively carried out, as shown in FIGURE~\ref{fig:depth_aware_pose}. 

\subsubsection{Depth-Aware Residual Pose Learning}
The final stage of our pipeline is a hierarchical pose refinement module that progressively updates pose estimates across pyramid levels. The pose refinement follows a residual learning scheme:

\begin{equation}
    \mathbf{T}_{i+1} = \mathbf{T}_i + \Delta \mathbf{T}_i,
\end{equation}
where $\mathbf{T}_i$ is the estimated pose at level $i$, and $\Delta \mathbf{T}_i$ is the residual correction predicted by our Depth-Aware Pose Network. Unlike traditional residual learning approaches that rely solely on image features, we introduce depth information as an additional cue to enhance translation accuracy.

\subsubsection{Depth-Aware Pose Estimation}  
Inspired by\cite{shan2020lio}, which demonstrated the benefits of integrating depth constraints in Lidar-Inertial Odometry, we explicitly incorporate depth cues into our pose refinement module. The input to our pose estimator consists of multi-scale optical flow features, depth information at the corresponding pyramid level, and previous pose estimates from coarser levels. We employ a multi-layer perceptron (MLP) to regress the residual pose update. The MLP consists of three fully connected layers of size (256,128,64) with Leaky-ReLU activations and dropout for regularization, followed by a final output layer to predict the pose update. The residual pose update is computed as:

\begin{equation}
    \Delta \mathbf{T}_i = \mathcal{P}(\mathbf{F}_i, D_i, \mathbf{T}_i),
\end{equation}
where $\mathcal{P}(\cdot)$ is the pose estimation network, $\mathbf{F}_i$ represents the extracted flow features, and $D_i$ is the depth map at level $i$. The depth term helps reduce scale ambiguity, particularly in low-texture regions.

\subsubsection{Uncertainty-Aware Pose Refinement}
To further improve robustness, we adopt an uncertainty-aware weighting mechanism\cite{teed2020raft}. The final pose estimate is computed as:

\begin{equation}
    \mathbf{T}_{final} = \sum_{i=0}^{L} w_i \mathbf{T}_i,
\end{equation}
where $w_i$ represents the confidence weight of each level’s pose prediction, and $L$ denotes the total number of pyramid levels. This strategy helps to mitigate the effect of noisy predictions from lower-resolution levels.

\begin{table*}[t]
\centering
\caption{Comparison with state-of-the-art Visual and LiDAR Odometry (VO/LO) methods on KITTI sequences 00-10. Our method D3LVO is trained on sequences 00-08 .The best results are bold, and the second best results are underlined.}
\label{tab:vo_lo_comparison}
\setlength{\tabcolsep}{1pt} 
\renewcommand{\arraystretch}{1.2} 
\begin{tabular}{l|cc|cc|cc|cc|cc|cc|cc|cc|cc|cc|cc|cc}
\hline
\multirow{2}{*}{\textbf{Method}} & 
\multicolumn{2}{c|}{\textbf{00}} & \multicolumn{2}{c|}{\textbf{01}} & \multicolumn{2}{c|}{\textbf{02}} & \multicolumn{2}{c|}{\textbf{03}} & \multicolumn{2}{c|}{\textbf{04}} & \multicolumn{2}{c|}{\textbf{05}} & \multicolumn{2}{c|}{\textbf{06}} & \multicolumn{2}{c|}{\textbf{07}} & \multicolumn{2}{c|}{\textbf{08}} & \multicolumn{2}{c|}{\textbf{09}} & \multicolumn{2}{c|}{\textbf{10}} & \multicolumn{2}{c|}{\textbf{Mean 07-10}} \\
& $t_{\text{rel}}$ & $r_{\text{rel}}$ & $t_{\text{rel}}$ & $r_{\text{rel}}$ 
& $t_{\text{rel}}$ & $r_{\text{rel}}$ & $t_{\text{rel}}$ & $r_{\text{rel}}$ 
& $t_{\text{rel}}$ & $r_{\text{rel}}$ & $t_{\text{rel}}$ & $r_{\text{rel}}$ 
& $t_{\text{rel}}$ & $r_{\text{rel}}$ & $t_{\text{rel}}$ & $r_{\text{rel}}$
& $t_{\text{rel}}$ & $r_{\text{rel}}$ & $t_{\text{rel}}$ & $r_{\text{rel}}$
& $t_{\text{rel}}$ & $r_{\text{rel}}$ & $t_{\text{rel}}$ & $r_{\text{rel}}$\\
\hline
\multicolumn{17}{l}{\textbf{Visual Odometry Methods:}} \\
\hline
SfMLearner~\cite{zhou2017unsupervised} & 21.32 & 6.19 & 22.41 & 2.79 & 24.10 & 4.18 & 12.56 & 4.52 & 4.32 & 3.28 & 12.99 & 4.66 & 15.55 & 5.58 & 12.61 & 6.31 & 10.66 & 3.75 & 11.32 & 4.07 & 15.52 & 4.06 & 12.46 & 4.55\\
DeepVO~\cite{wang2017deepvo} & - & - & - & - & - & - & 8.49 & 6.89 & 7.19 & 6.97 & 2.62 & 3.61 & 5.42 & 5.82 & 3.19 & 4.60 & - & - & - & - & 8.11 & 8.83 & 6.01 & 6.72\\
DFVO~\cite{zhan2021df}  & 2.25 & 0.58 & 66.98 & 17.04 & 3.60 & 0.52 & 2.67 & 0.50 & 1.43 & \underline{0.29} & 1.10 & \textbf{0.30} & 1.03 & \textbf{0.30} & 0.97 & \textbf{0.27} & 1.60 & \textbf{0.32} & 2.61 & \textbf{0.29} & 2.29 & \underline{0.37} & 1.87 & \textbf{0.31}\\
Li et al.~\cite{li2021generalizing} & 1.32 & 0.45 & 2.83 & 0.65 & \underline{1.42} & \textbf{0.45} & 1.77 & \underline{0.39} & 1.22 & \textbf{0.27} & 1.07 & 0.44 & 1.02 & 0.41 & 2.06 & 1.18 & 1.50 & 0.42 & 1.87 & 0.46 & 1.93 & \textbf{0.30} & 1.84 & 0.59 \\
\hline
\multicolumn{17}{l}{\textbf{LiDAR Odometry Methods:}} \\
\hline
DeepLO~\cite{cho2019deeplo} & 1.90 & 0.80 & 37.83 & 0.86 & 2.05 & 0.81 & 2.85 & 1.43 & 1.54 & 0.87 & 1.72 & 0.92 & 0.84 & 0.47 & \textbf{0.70} & 0.67 & 1.81 & 1.02 & 6.55 & 2.19 & 7.74 & 2.84 & 4.20 & 1.68 \\
LO-Net~\cite{li2019net} & 1.47 & 0.72 & 1.36 & 0.47 & 1.52 & 0.71 & \underline{1.03} & 0.66 & \underline{0.51} & 0.65 & 1.04 & 0.69 & 0.71 & 0.50 & 1.70 & 0.89 & 2.12 & 0.77 & 1.37 & 0.58 & 1.80 & 0.93 & 1.75 & 0.79 \\
PWCLO~\cite{wang2021pwclo} & \underline{0.89} & \underline{0.43} & 1.11 & \underline{0.42} & 1.87 & 0.76 & 1.42 & 0.92 & 1.15 & 0.94 & 1.34 & 0.71 & \underline{0.60} & 0.38 & 1.16 & 1.00 & 1.68 & 0.72 & \textbf{0.88} & 0.46 & 2.14 & 0.71 & 1.47 & 0.72 \\
LodoNet~\cite{zheng2020lodonet} & 1.43 & 0.69 & \underline{0.96} & \textbf{0.28} & 1.46 & 0.57 & 2.12 & 0.98 & 0.65 & 0.45 & 1.07 & 0.59 & 0.62 & 0.34 & 1.86 & 1.64 & 2.04 & 0.97 & 0.63 & 0.35 & 1.18 & 0.45 & 1.43 & 0.85\\
\hline
\multicolumn{17}{l}{\textbf{multimodal Odometry Methods:}} \\
\hline
An et al.~\cite{an2022visual} & 2.53 & 0.79 & 3.76 & 0.80 & 3.95 & 1.05 & 2.75 & 1.39 & 1.81 & 1.48 & 3.49 & 0.79 & 1.84 & 0.83 & 3.27 & 1.51 & 2.75 & 1.61 & 3.70 & 1.83 & 4.65 & 0.51 & 3.59 & 1.37 \\
H-VLO~\cite{aydemir2022h} & 1.75 & 0.62 & 43.2 & 0.46 & 2.32 & 0.60 & 2.52 & 0.47 & 0.73 & 0.36 & \underline{0.85} & 0.35 & 0.75 & \textbf{0.30} & \underline{0.79} & 0.48 & \underline{1.35} & 0.38 & 1.89 & \underline{0.34} & \underline{1.39} & 0.52 & \underline{1.36} & 0.43 \\
\textbf{Ours} & \textbf{0.88} & \textbf{0.41} & \textbf{0.91} & 0.45 & \textbf{1.32} & \underline{0.47} & \textbf{0.87} & \textbf{0.35} & \textbf{0.36} & 0.52 & \textbf{0.77} & \textbf{0.22} & \textbf{0.54} & \underline{0.32} & 0.82 & \underline{0.30} & \textbf{1.12} & \underline{0.36} & \underline{0.97} & 0.41 & \textbf{0.95} & 0.47 & \textbf{0.97} & \underline{0.39}\\
\hline
\end{tabular}
\end{table*}

\begin{table*}[t]
\centering
\caption{Comparison with traditional Visual-LiDAR odometry methods on KITTI sequences 00-10. Our method D3LVO is trained on sequences 00-08. The best results are bold, and the second best results are underlined.}
\label{tab:traditional_vlo_comparison}
\resizebox{\textwidth}{!}{%
\begin{tabular}{l|ccccccccccc|c}
\hline
& \textbf{00} & \textbf{01} & \textbf{02} & \textbf{03} & \textbf{04} & \textbf{05} & \textbf{06} & \textbf{07} & \textbf{08} & \textbf{09} & \textbf{10} & \textbf{Mean 00-10} \\
\textbf{Method} & $t_{\text{rel}}$ & $t_{\text{rel}}$ & $t_{\text{rel}}$ & $t_{\text{rel}}$ & $t_{\text{rel}}$ & $t_{\text{rel}}$ & $t_{\text{rel}}$ & $t_{\text{rel}}$ & $t_{\text{rel}}$ & $t_{\text{rel}}$ & $t_{\text{rel}}$ & $t_{\text{rel}}$ \\
\hline
LeGO-LOAM~\cite{shan2018lego} & 1.51 & - & 1.96 & 1.41 & 1.69 & 1.01 & 0.90 & \underline{0.81} & 1.48 & 1.57 & 1.81 & 1.42 \\
DVL-SLAM~\cite{shin2020dvl} & \underline{0.93} & \underline{1.47} & \textbf{1.11} & 0.92 & 0.67 & 0.82 & 0.92 & 1.26 & 1.32 & \textbf{0.66} & \textbf{0.70} & 0.98 \\
PL-LOAM~\cite{huang2020lidar} & 0.99 & 1.87 & 1.38 & \textbf{0.65} & \underline{0.42} & \textbf{0.72} & \underline{0.61} & \textbf{0.56} & \underline{1.27} & 1.06 & \underline{0.83} & \underline{0.94} \\
\textbf{Ours} & \textbf{0.88} & \textbf{0.91} & \underline{1.32} & \underline{0.87} & \textbf{0.36} & \underline{0.77} & \textbf{0.54} & 0.82 & \textbf{1.12} & \underline{0.97} & 0.95 & \textbf{0.86} \\
\hline
\end{tabular}}
\end{table*}

\subsection{Loss Function}

To supervise pose estimation, we adopt a scale-aware loss inspired by\cite{wang2022efficient}, which introduces learnable scale parameters to balance translation and rotation components effectively. The loss at the $l$-th pyramid level is defined as:
\begin{equation}
    \begin{split}
        \ell^l = & 
        \left\| \mathbf{t}_{\text{gt}} - \mathbf{t}^l \right\|_1 \exp(-s_t) + s_t 
        + \\
        & \left\| \frac{\mathbf{q}_{\text{gt}}}{\|\mathbf{q}_{\text{gt}}\|_2} - \frac{\mathbf{q}^l}{\|\mathbf{q}^l\|_2} \right\|_2^2 \exp(-s_q) + s_q,
        \label{eq:loss_function}
    \end{split}
\end{equation}
where $\mathbf{t}_{\text{gt}}$ and $\mathbf{q}_{\text{gt}}$ are the ground-truth translation vector and quaternion, respectively, while $\mathbf{t}^l$ and $\mathbf{q}^l$ represent the predicted translation and quaternion at the $l$-th level. The terms $s_t$ and $s_q$ are learnable parameters used to balance the translation and rotation components, adapting to their varying scales and units.

The overall loss $\ell$, which aggregates multi-scale supervision is defined as:
\begin{equation}
    \ell = \sum_{l=1}^L \alpha^l \ell^l,
\end{equation}
where $L$ denotes the total number of pyramid levels, and $\alpha^l$ is a hyper-parameter that weights the contribution of each level. This formulation ensures robust supervision across hierarchical scales, addressing the challenges posed by diverse scene geometries and motion patterns.

\section{Experiments}
\subsection{KITTI Odometry Dataset}
We evaluate our method on the KITTI odometry benchmark\cite{geiger2012we}, a widely used dataset for assessing odometry performance in real-world driving scenarios. The dataset provides synchronized stereo RGB images, sparse LiDAR point clouds, and ground-truth poses from GPS/IMU. We focus on sequences 00–10, which include urban, highway, and countryside driving conditions. To enhance the depth information, we utilize the depth completion dataset, where sparse LiDAR scans are converted into dense depth maps, improving the robustness of feature extraction and motion estimation.

\subsection{Implementation Details}
Our model is implemented using PyTorch 1.12.1 and trained on an NVIDIA RTX 3080 Ti GPU. The input images are resized to a fixed resolution of $1216\times352$ to maintain consistency across sequences. We use the sparse LiDAR depth maps from the KITTI depth completion dataset and the corresponding RGB images from the KITTI Odometry dataset as inputs. Data augmentation includes random horizontal flipping, brightness adjustment, and slight rotation perturbations to improve generalization.

\subsection{Evaluation Metrics}

We evaluate our method using two metrics:

\begin{itemize}
    \item \textbf{Translational RMSE (\%):} Measures the root mean squared error of the translation for each sequence relative to the ground truth.
    \item \textbf{Rotational RMSE ($^\circ$/100m):} Measures the average angular error per 100 meters of travel.
\end{itemize}

The average translational RMSE (\%) and the average rotational RMSE ($^\circ$/100m) are calculated over the 00-10 subsequences with lengths of 100, 200, ..., 800m  in accordance with the standard odometry benchmark protocol\cite{wang2021pwclo}. 

\subsection{Quantitative Results} 
We conducted extensive evaluations on the KITTI odometry dataset (sequences 00-10) and compared our method with state-of-the-art odometry approaches. Across the section, average translational RMSE and average rotational RMSE are denoted as $t_{\text{rel}}$ and $r_{\text{rel}}$ respective. To assess the effectiveness of our approach, we present comparisons in three categories: single sensor (visual or LiDAR) based odometry methods, traditional multimodal odometry methods, and learning-based multimodal odometry methods.

\subsubsection{Comparison with Visual or LiDAR Odometry Methods}
We compare our approach with various odometry methods with single sensor input, primarily visual odometry, LiDAR odometry methods for relevance. As shown in TABLE~\ref{tab:vo_lo_comparison}, our method achieves competitive performance across multiple evaluation metrics.

The quantitative results indicate that our algorithm, trained only on sequences 00-08, demonstrates superior performance on sequences 09-10 as well. This suggests that our method generalizes well to unseen data, which is crucial for real-world applications.

\subsubsection{Comparison with Traditional Multimodal Odometry Methods}
We compare our approach with several well-known traditional multimodal odometry methods, including LeGO-LOAM~\cite{shan2018lego}, DVL-SLAM~\cite{shin2020dvl}, and PL-LOAM~\cite{huang2020lidar}. As shown in TABLE~\ref{tab:traditional_vlo_comparison}, our method achieves the lowest RMSE values in five out of the 11 sequences and outperforms all baselines in terms of average RMSE. These results demonstrate the effectiveness of our approach in integrating multimodal data for accurate odometry estimation. Compared to PL-LOAM\cite{huang2020lidar}, our method has a 8.5\% decline in the mean translation error ($t_{\text{rel}}$) on sequence 00-10 .

\subsubsection{Comparison with Learning-Based Multimodal Odometry Methods} 
We also compare our method to learning-based multimodal odometry approaches, such as Self-VLO\cite{li2021self},SelfVIO\cite{almalioglu2022selfvio} etc. . These methods leverage different feature representations and specifically designed neural networks to extract and fuse features for odometry estimation. TABLE~\ref{tab:vo_lo_comparison} and TABLE~\ref{tab:learning_based_vlo} shows that our approach consistently outperforms most of the learning-based methods in both translational and rotational RMSE, demonstrating the advantages of our dense depth feature representation for pose estimation. For example, our method achieves an 73\% lower mean translation error $t_{\text{rel}}$ and a 72\% lower rotational error $r_{\text{rel}}$ on sequences 07 and 10 compared to the An et al.\cite{an2022visual}.

\begin{table}[h]
\centering
\caption{Comparison with learning-based multimodal odometry methods on  KITTI sequences 09-10. The best results for each sequence are \textbf{bold}, and the second best results are underlined.}
\label{tab:learning_based_vlo}
\scalebox{0.85}{ 
\begin{tabular}{l|c|cc|cc|cc}
\hline
\multirow{2}{*}{\textbf{Method}} & 
\multirow{2}{*}{\textbf{Modalities}} & 
\multicolumn{2}{c|}{\textbf{09}} & \multicolumn{2}{c|}{\textbf{10}} & \multicolumn{2}{c|}{\textbf{Mean 09-10}} \\
& & $t_{\text{rel}}$ & $r_{\text{rel}}$ & $t_{\text{rel}}$ & $r_{\text{rel}}$ & $t_{\text{rel}}$ & $r_{\text{rel}}$ \\
\hline
Self-VLO~\cite{li2021self} & visual+LiDAR & 2.58 & 1.13 & 2.67 & 1.28 & 2.62 & 1.21 \\
SelfVIO~\cite{almalioglu2022selfvio} & visual+inertial & 1.95 & 1.15 & 1.81 & 1.30 & 1.88 & 1.23 \\
VIOLearner~\cite{shamwell2019unsupervised} & visual+inertial & \underline{1.82} & 1.08 & 1.74 & 1.38 & 1.78 & 1.23\\
H-VLO~\cite{aydemir2022h} & visual+LiDAR & 1.89 & \textbf{0.34} & \underline{1.39} & \underline{0.52} & \underline{1.67} & \textbf{0.43}\\
\textbf{Ours} & visual+LiDAR & \textbf{0.97} & \underline{0.41} & \textbf{0.95} & \textbf{0.47} & \textbf{0.96} & \underline{0.44}\\
\hline
\end{tabular}}
\end{table}

\begin{table}[h]
\centering
\caption{Ablation study on the effect of depth information on test KITTI sequences 09-10.The best results for each sequence are bold.}
\label{tab:ablation_study}
\scalebox{0.90}{ 
\begin{tabular}{l|cc|cc|cc}
\hline
& \multicolumn{2}{c|}{\textbf{09}} & \multicolumn{2}{c|}{\textbf{10}} & \multicolumn{2}{c}{\textbf{Mean 09-10}} \\
\textbf{Configuration} & $t_{\text{rel}}$ & $r_{\text{rel}}$ & $t_{\text{rel}}$ & $r_{\text{rel}}$ & $t_{\text{rel}}$ & $r_{\text{rel}}$ \\
\hline
RGB-only & 10.37 & 1.84 & 5.65 & 2.35 & 8.01 & 2.10 \\
RGB + Sparse Depth & 2.78 & 0.72 & 3.73 & 0.91 & 3.26 & 0.82 \\
\textbf{RGB + Depth Completion} & \textbf{0.97} & \textbf{0.41} & \textbf{0.95} & \textbf{0.47} & \textbf{0.96} & \textbf{0.44} \\
\hline
\end{tabular}}
\end{table}

\subsection{Ablation Study and Visualization}
To further investigate the impact of depth information on pose estimation, we conduct an ablation study on KITTI sequences 09-10. We compare three configurations:

\begin{itemize}
    \item RGB-only: Uses only RGB images for feature extraction and pose estimation.
    \item RGB + Sparse Depth: Uses raw sparse depth points and fills the empty pixels' depth with a default value.
    \item RGB + Depth Completion (Ours): Incorporates completed depth maps to enhance depth-aware optical flow prediction and feature fusion.
\end{itemize}
As shown in TABLE~\ref{tab:ablation_study}, removing depth information leads to a significant degradation in translational and rotational accuracy. Notably, our dense-depth based method achieves the lowest errors, demonstrating that depth completion effectively reduces depth ambiguity and improves pose estimation robustness. The results demonstrate that utilizing dense depth maps significantly improves performance compared to only RGB images or sparse depth configuration and validate the effectiveness of our depth-aware approach. Under the average of sequences 09 and 10, our dense-depth based method achieves an 88\% lower mean translation error $t_{\text{rel}}$ and a 79\% lower rotational error $r_{\text{rel}}$ compared to the RGB-only method.

\begin{figure}[htbp] \centering \includegraphics[width=0.48\textwidth]{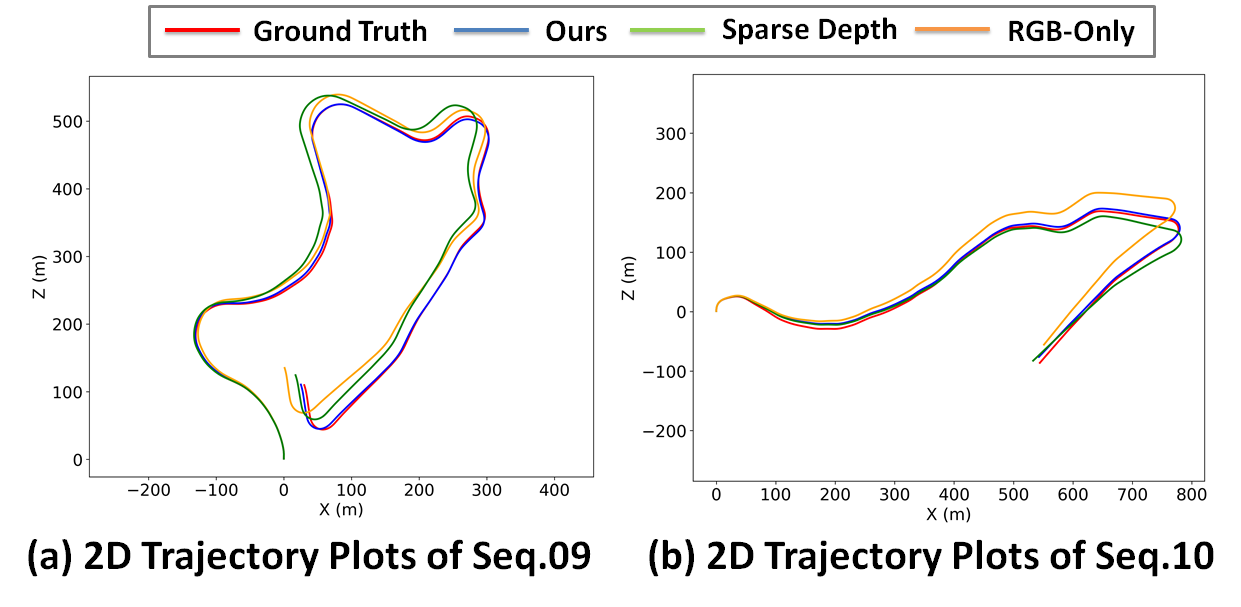} 
\caption{2D trajectory comparison on KITTI sequences 09 and 10. The dense-depth based approach shows better alignment with the ground truth, particularly in challenging motion scenarios.} \label{fig:traj_comparison} 
\end{figure}

To provide qualitative insights into the performance of our method, we visualize the 2D trajectory results for KITTI sequences 09 and 10 in FIGURE~\ref{fig:traj_comparison}. We compare our current approach with RGB-only input and RGB with sparse depth from raw LiDAR signal. As shown in FIGURE~\ref{fig:traj_comparison}, the trajectory predicted with the current approach (RGB+depth Completion) aligns better with the ground truth trajectory compared to the other two approaches, especially in challenging turns where rotational angle estimation may be unreliable.

\section{Conclusion}
In this paper, we proposed a novel Lidar-Visual Odometry framework that utilizes dense depth maps to enhance feature extraction and guide optical flow estimation, resulting in more accurate pose estimation. 
We utilize depth completion to reduce depth ambiguity in textureless regions, and improve feature quality through depth-aware feature fusion. By leveraging depth-aware flow prediction and hierarchical pose refinement, our method achieves superior pose accuracy, outperforming most image/LiDAR and multimodal odometry methods in the KITTI benchmark. While our method relies on the performance of depth-completion, these methods are well developed with good generalization ability\cite{zhou2017unsupervised,hu2021penet,song2021self,xie2022recent}. Future work will evaluate the method's generalization ability across diverse datasets and integration into SLAM systems. Other possible improvements include building an end-to-end system with multi-task approaches and self-supervising \cite{wan2022multi}, extending the method's applicability in real world scenarios.

\section*{Acknowledgments}
The work was supported by Guangdong Provincial Quantum Science Strategic Initiative
(GDZX2306001) and the startup fund of Shenzhen City.

\bibliographystyle{ieeetran}
\bibliography{references}

\begin{thebibliography}{10}
\providecommand{\url}[1]{#1}
\csname url@samestyle\endcsname
\providecommand{\newblock}{\relax}
\providecommand{\bibinfo}[2]{#2}
\providecommand{\BIBentrySTDinterwordspacing}{\spaceskip=0pt\relax}
\providecommand{\BIBentryALTinterwordstretchfactor}{4}
\providecommand{\BIBentryALTinterwordspacing}{\spaceskip=\fontdimen2\font plus
\BIBentryALTinterwordstretchfactor\fontdimen3\font minus
  \fontdimen4\font\relax}
\providecommand{\BIBforeignlanguage}[2]{{%
\expandafter\ifx\csname l@#1\endcsname\relax
\typeout{** WARNING: IEEEtran.bst: No hyphenation pattern has been}%
\typeout{** loaded for the language `#1'. Using the pattern for}%
\typeout{** the default language instead.}%
\else
\language=\csname l@#1\endcsname
\fi
#2}}
\providecommand{\BIBdecl}{\relax}
\BIBdecl

\bibitem{shan2020lio}
T.~Shan, B.~Englot, D.~Meyers, W.~Wang, C.~Ratti, and D.~Rus, ``Lio-sam:
  Tightly-coupled lidar inertial odometry via smoothing and mapping,'' in
  \emph{2020 IEEE/RSJ international conference on intelligent robots and
  systems (IROS)}.\hskip 1em plus 0.5em minus 0.4em\relax IEEE, 2020, pp.
  5135--5142.

\bibitem{zhang2015visual}
J.~Zhang and S.~Singh, ``Visual-lidar odometry and mapping: Low-drift, robust,
  and fast,'' in \emph{2015 IEEE international conference on robotics and
  automation (ICRA)}.\hskip 1em plus 0.5em minus 0.4em\relax IEEE, 2015, pp.
  2174--2181.

\bibitem{li2021self}
B.~Li, M.~Hu, S.~Wang, L.~Wang, and X.~Gong, ``Self-supervised visual-lidar
  odometry with flip consistency,'' in \emph{Proceedings of the IEEE/CVF Winter
  Conference on Applications of Computer Vision}, 2021, pp. 3844--3852.

\bibitem{tateno2017cnn}
K.~Tateno, F.~Tombari, I.~Laina, and N.~Navab, ``Cnn-slam: Real-time dense
  monocular slam with learned depth prediction,'' in \emph{Proceedings of the
  IEEE conference on computer vision and pattern recognition}, 2017, pp.
  6243--6252.

\bibitem{sun2018pwc}
D.~Sun, X.~Yang, M.-Y. Liu, and J.~Kautz, ``Pwc-net: Cnns for optical flow
  using pyramid, warping, and cost volume,'' in \emph{Proceedings of the IEEE
  conference on computer vision and pattern recognition}, 2018, pp. 8934--8943.

\bibitem{du2021lidar}
S.~Du, Y.~Li, X.~Li, and M.~Wu, ``Lidar odometry and mapping based on semantic
  information for outdoor environment,'' \emph{Remote Sensing}, vol.~13,
  no.~15, p. 2864, 2021.

\bibitem{zhang2014loam}
J.~Zhang, S.~Singh \emph{et~al.}, ``Loam: Lidar odometry and mapping in
  real-time.'' in \emph{Robotics: Science and systems}, vol.~2, no.~9.\hskip
  1em plus 0.5em minus 0.4em\relax Berkeley, CA, 2014, pp. 1--9.

\bibitem{shan2018lego}
T.~Shan and B.~Englot, ``Lego-loam: Lightweight and ground-optimized lidar
  odometry and mapping on variable terrain,'' in \emph{2018 IEEE/RSJ
  International Conference on Intelligent Robots and Systems (IROS)}.\hskip 1em
  plus 0.5em minus 0.4em\relax IEEE, 2018, pp. 4758--4765.

\bibitem{geiger2013vision}
A.~Geiger, P.~Lenz, C.~Stiller, and R.~Urtasun, ``Vision meets robotics: The
  kitti dataset,'' \emph{The International Journal of Robotics Research},
  vol.~32, no.~11, pp. 1231--1237, 2013.

\bibitem{xu2019depth}
Y.~Xu, X.~Zhu, J.~Shi, G.~Zhang, H.~Bao, and H.~Li, ``Depth completion from
  sparse lidar data with depth-normal constraints,'' in \emph{Proceedings of
  the IEEE/CVF International Conference on Computer Vision}, 2019, pp.
  2811--2820.

\bibitem{vaswani2017attention}
A.~Vaswani, ``Attention is all you need,'' \emph{Advances in Neural Information
  Processing Systems}, 2017.

\bibitem{geiger2012we}
A.~Geiger, P.~Lenz, and R.~Urtasun, ``Are we ready for autonomous driving? the
  kitti vision benchmark suite,'' in \emph{2012 IEEE conference on computer
  vision and pattern recognition}.\hskip 1em plus 0.5em minus 0.4em\relax IEEE,
  2012, pp. 3354--3361.

\bibitem{mur2015orb}
R.~Mur-Artal, J.~M.~M. Montiel, and J.~D. Tardos, ``Orb-slam: a versatile and
  accurate monocular slam system,'' \emph{IEEE transactions on robotics},
  vol.~31, no.~5, pp. 1147--1163, 2015.

\bibitem{mur2017orb}
R.~Mur-Artal and J.~D. Tard{\'o}s, ``Orb-slam2: An open-source slam system for
  monocular, stereo, and rgb-d cameras,'' \emph{IEEE transactions on robotics},
  vol.~33, no.~5, pp. 1255--1262, 2017.

\bibitem{engel2017direct}
J.~Engel, V.~Koltun, and D.~Cremers, ``Direct sparse odometry,'' \emph{IEEE
  transactions on pattern analysis and machine intelligence}, vol.~40, no.~3,
  pp. 611--625, 2017.

\bibitem{newcombe2011dtam}
R.~A. Newcombe, S.~J. Lovegrove, and A.~J. Davison, ``Dtam: Dense tracking and
  mapping in real-time,'' in \emph{2011 international conference on computer
  vision}.\hskip 1em plus 0.5em minus 0.4em\relax IEEE, 2011, pp. 2320--2327.

\bibitem{engel2014lsd}
J.~Engel, T.~Sch{\"o}ps, and D.~Cremers, ``Lsd-slam: Large-scale direct
  monocular slam,'' in \emph{European conference on computer vision}.\hskip 1em
  plus 0.5em minus 0.4em\relax Springer, 2014, pp. 834--849.

\bibitem{wang2017deepvo}
S.~Wang, R.~Clark, H.~Wen, and N.~Trigoni, ``Deepvo: Towards end-to-end visual
  odometry with deep recurrent convolutional neural networks,'' in \emph{2017
  IEEE international conference on robotics and automation (ICRA)}.\hskip 1em
  plus 0.5em minus 0.4em\relax IEEE, 2017, pp. 2043--2050.

\bibitem{zhan2021df}
H.~Zhan, C.~S. Weerasekera, J.-W. Bian, R.~Garg, and I.~Reid, ``Df-vo: What
  should be learnt for visual odometry?'' \emph{arXiv preprint
  arXiv:2103.00933}, 2021.

\bibitem{teed2020raft}
Z.~Teed and J.~Deng, ``Raft: Recurrent all-pairs field transforms for optical
  flow,'' in \emph{Computer Vision--ECCV 2020: 16th European Conference,
  Glasgow, UK, August 23--28, 2020, Proceedings, Part II 16}.\hskip 1em plus
  0.5em minus 0.4em\relax Springer, 2020, pp. 402--419.

\bibitem{zhan2018unsupervised}
H.~Zhan, R.~Garg, C.~S. Weerasekera, K.~Li, H.~Agarwal, and I.~Reid,
  ``Unsupervised learning of monocular depth estimation and visual odometry
  with deep feature reconstruction,'' in \emph{Proceedings of the IEEE
  conference on computer vision and pattern recognition}, 2018, pp. 340--349.

\bibitem{yang2020d3vo}
N.~Yang, L.~v. Stumberg, R.~Wang, and D.~Cremers, ``D3vo: Deep depth, deep pose
  and deep uncertainty for monocular visual odometry,'' in \emph{Proceedings of
  the IEEE/CVF conference on computer vision and pattern recognition}, 2020,
  pp. 1281--1292.

\bibitem{besl1992method}
P.~J. Besl and N.~D. McKay, ``Method for registration of 3-d shapes,'' in
  \emph{Sensor fusion IV: control paradigms and data structures}, vol.
  1611.\hskip 1em plus 0.5em minus 0.4em\relax Spie, 1992, pp. 586--606.

\bibitem{ye2019tightly}
H.~Ye, Y.~Chen, and M.~Liu, ``Tightly coupled 3d lidar inertial odometry and
  mapping,'' in \emph{2019 International Conference on Robotics and Automation
  (ICRA)}.\hskip 1em plus 0.5em minus 0.4em\relax IEEE, 2019, pp. 3144--3150.

\bibitem{li2019net}
Q.~Li, S.~Chen, C.~Wang, X.~Li, C.~Wen, M.~Cheng, and J.~Li, ``Lo-net: Deep
  real-time lidar odometry,'' in \emph{Proceedings of the IEEE/CVF Conference
  on Computer Vision and Pattern Recognition}, 2019, pp. 8473--8482.

\bibitem{cho2019deeplo}
Y.~Cho, G.~Kim, and A.~Kim, ``Deeplo: Geometry-aware deep lidar odometry,''
  \emph{arXiv preprint arXiv:1902.10562}, 2019.

\bibitem{wang2021pwclo}
G.~Wang, X.~Wu, Z.~Liu, and H.~Wang, ``Pwclo-net: Deep lidar odometry in 3d
  point clouds using hierarchical embedding mask optimization,'' in
  \emph{Proceedings of the IEEE/CVF conference on computer vision and pattern
  recognition}, 2021, pp. 15\,910--15\,919.

\bibitem{zheng2020lodonet}
C.~Zheng, Y.~Lyu, M.~Li, and Z.~Zhang, ``Lodonet: A deep neural network with 2d
  keypoint matching for 3d lidar odometry estimation,'' in \emph{Proceedings of
  the 28th ACM international conference on multimedia}, 2020, pp. 2391--2399.

\bibitem{maddern2014illumination}
W.~Maddern, A.~Stewart, C.~McManus, B.~Upcroft, W.~Churchill, and P.~Newman,
  ``Illumination invariant imaging: Applications in robust vision-based
  localisation, mapping and classification for autonomous vehicles,'' in
  \emph{Proceedings of the Visual Place Recognition in Changing Environments
  Workshop, IEEE International Conference on Robotics and Automation (ICRA),
  Hong Kong, China}, vol.~2, no.~3, 2014, p.~5.

\bibitem{liu2024dvlo}
J.~Liu, D.~Zhuo, Z.~Feng, S.~Zhu, C.~Peng, Z.~Liu, and H.~Wang, ``Dvlo: Deep
  visual-lidar odometry with local-to-global feature fusion and bi-directional
  structure alignment,'' in \emph{European Conference on Computer
  Vision}.\hskip 1em plus 0.5em minus 0.4em\relax Springer, 2024, pp. 475--493.

\bibitem{an2022visual}
Y.~An, J.~Shi, D.~Gu, and Q.~Liu, ``Visual-lidar slam based on unsupervised
  multi-channel deep neural networks,'' \emph{Cognitive Computation}, vol.~14,
  no.~4, pp. 1496--1508, 2022.

\bibitem{hu2021penet}
M.~Hu, S.~Wang, B.~Li, S.~Ning, L.~Fan, and X.~Gong, ``Penet: Towards precise
  and efficient image guided depth completion,'' in \emph{2021 IEEE
  International Conference on Robotics and Automation (ICRA)}.\hskip 1em plus
  0.5em minus 0.4em\relax IEEE, 2021, pp. 13\,656--13\,662.

\bibitem{hu2018squeeze}
J.~Hu, L.~Shen, and G.~Sun, ``Squeeze-and-excitation networks,'' in
  \emph{Proceedings of the IEEE conference on computer vision and pattern
  recognition}, 2018, pp. 7132--7141.

\bibitem{woo2018cbam}
S.~Woo, J.~Park, J.-Y. Lee, and I.~S. Kweon, ``Cbam: Convolutional block
  attention module,'' in \emph{Proceedings of the European conference on
  computer vision (ECCV)}, 2018, pp. 3--19.

\bibitem{hui2018liteflownet}
T.-W. Hui, X.~Tang, and C.~C. Loy, ``Liteflownet: A lightweight convolutional
  neural network for optical flow estimation,'' in \emph{Proceedings of the
  IEEE conference on computer vision and pattern recognition}, 2018, pp.
  8981--8989.

\bibitem{zhao2020maskflownet}
S.~Zhao, Y.~Sheng, Y.~Dong, E.~I. Chang, Y.~Xu \emph{et~al.}, ``Maskflownet:
  Asymmetric feature matching with learnable occlusion mask,'' in
  \emph{Proceedings of the IEEE/CVF conference on computer vision and pattern
  recognition}, 2020, pp. 6278--6287.

\bibitem{mittal2020just}
H.~Mittal, B.~Okorn, and D.~Held, ``Just go with the flow: Self-supervised
  scene flow estimation,'' in \emph{Proceedings of the IEEE/CVF conference on
  computer vision and pattern recognition}, 2020, pp. 11\,177--11\,185.

\bibitem{dong2023rethinking}
Q.~Dong, C.~Cao, and Y.~Fu, ``Rethinking optical flow from geometric matching
  consistent perspective,'' in \emph{Proceedings of the IEEE/CVF Conference on
  Computer Vision and Pattern Recognition}, 2023, pp. 1337--1347.

\bibitem{bao2019depth}
W.~Bao, W.-S. Lai, C.~Ma, X.~Zhang, Z.~Gao, and M.-H. Yang, ``Depth-aware video
  frame interpolation,'' in \emph{Proceedings of the IEEE/CVF conference on
  computer vision and pattern recognition}, 2019, pp. 3703--3712.

\bibitem{zhou2017unsupervised}
T.~Zhou, M.~Brown, N.~Snavely, and D.~G. Lowe, ``Unsupervised learning of depth
  and ego-motion from video,'' in \emph{Proceedings of the IEEE conference on
  computer vision and pattern recognition}, 2017, pp. 1851--1858.

\bibitem{li2021generalizing}
S.~Li, X.~Wu, Y.~Cao, and H.~Zha, ``Generalizing to the open world: Deep visual
  odometry with online adaptation,'' in \emph{Proceedings of the IEEE/CVF
  Conference on Computer Vision and Pattern Recognition}, 2021, pp.
  13\,184--13\,193.

\bibitem{aydemir2022h}
E.~Aydemir, N.~Fetic, and M.~Unel, ``H-vlo: hybrid lidar-camera fusion for
  self-supervised odometry,'' in \emph{2022 IEEE/RSJ international conference
  on intelligent robots and systems (IROS)}.\hskip 1em plus 0.5em minus
  0.4em\relax IEEE, 2022, pp. 3302--3307.

\bibitem{shin2020dvl}
Y.-S. Shin, Y.~S. Park, and A.~Kim, ``Dvl-slam: Sparse depth enhanced direct
  visual-lidar slam,'' \emph{Autonomous Robots}, vol.~44, no.~2, pp. 115--130,
  2020.

\bibitem{huang2020lidar}
S.-S. Huang, Z.-Y. Ma, T.-J. Mu, H.~Fu, and S.-M. Hu, ``Lidar-monocular visual
  odometry using point and line features,'' in \emph{2020 IEEE international
  conference on robotics and automation (ICRA)}.\hskip 1em plus 0.5em minus
  0.4em\relax IEEE, 2020, pp. 1091--1097.

\bibitem{wang2022efficient}
G.~Wang, X.~Wu, S.~Jiang, Z.~Liu, and H.~Wang, ``Efficient 3d deep lidar
  odometry,'' \emph{IEEE Transactions on Pattern Analysis and Machine
  Intelligence}, vol.~45, no.~5, pp. 5749--5765, 2022.

\bibitem{almalioglu2022selfvio}
Y.~Almalioglu, M.~Turan, M.~R.~U. Saputra, P.~P. De~Gusm{\~a}o, A.~Markham, and
  N.~Trigoni, ``Selfvio: Self-supervised deep monocular visual--inertial
  odometry and depth estimation,'' \emph{Neural Networks}, vol. 150, pp.
  119--136, 2022.

\bibitem{shamwell2019unsupervised}
E.~J. Shamwell, K.~Lindgren, S.~Leung, and W.~D. Nothwang, ``Unsupervised deep
  visual-inertial odometry with online error correction for rgb-d imagery,''
  \emph{IEEE transactions on pattern analysis and machine intelligence},
  vol.~42, no.~10, pp. 2478--2493, 2019.

\bibitem{song2021self}
Z.~Song, J.~Lu, Y.~Yao, and J.~Zhang, ``Self-supervised depth completion from
  direct visual-lidar odometry in autonomous driving,'' \emph{IEEE Transactions
  on Intelligent Transportation Systems}, vol.~23, no.~8, pp. 11\,654--11\,665,
  2021.

\bibitem{xie2022recent}
Z.~Xie, X.~Yu, X.~Gao, K.~Li, and S.~Shen, ``Recent advances in conventional
  and deep learning-based depth completion: A survey,'' \emph{IEEE Transactions
  on Neural Networks and Learning Systems}, vol.~35, no.~3, pp. 3395--3415,
  2022.

\bibitem{wan2022multi}
Y.~Wan, Q.~Zhao, C.~Guo, C.~Xu, and L.~Fang, ``Multi-sensor fusion
  self-supervised deep odometry and depth estimation,'' \emph{Remote Sensing},
  vol.~14, no.~5, p. 1228, 2022.

\end{thebibliography}
 
\vspace{11pt}

\vfill

\end{document}